\let\oldvec\vec
\documentclass[runningheads,a4paper]{llncs}
\let\vec\oldvec

\usepackage{amssymb}
\usepackage{graphicx}

\usepackage[utf8]{inputenc}
\usepackage{amsmath}
\usepackage{mathtools}

\usepackage[normalem]{ulem}

\usepackage{algorithm}
\usepackage[noend]{algpseudocode}

\graphicspath{{./gfx/}}
\DeclareGraphicsExtensions{.eps}

\usepackage{tikz}
\usetikzlibrary{shapes}

\usepackage[caption=false]{subfig}

\usepackage{booktabs}
\usepackage{siunitx}

\DeclareMathOperator*{\argmax}{arg\,max}
\DeclareMathOperator{\Lagr}{\mathcal{L}}

\DeclarePairedDelimiterX{\ExpArg}[1]{[}{]}{#1}

\newcommand\indep{\protect\mathpalette{\protect\independenT}{\perp}}
\def\independenT#1#2{\mathrel{\rlap{$#1#2$}\mkern2mu{#1#2}}}

\usepackage{hyperref}
\usepackage[
    backend=biber,
    hyperref=true,
    style=numeric,
    maxbibnames=99, 
    sorting=nyt,
    firstinits=true,
    uniquename=init,
    isbn=false,
    url=false,
    doi=false,
    eprint=false
]{biblatex}
\usepackage{url}
\urldef{\mailsa}\path|{maxime.gasse, alexandre.aussem}@liris.cnrs.fr|    
\newcommand{\keywords}[1]{\par\addvspace\baselineskip
\noindent\keywordname\enspace\ignorespaces#1}

\begin{document}

\mainmatter  

\title{F-measure Maximization in Multi-Label Classification with Conditionally Independent Label Subsets}

\titlerunning{F-measure Maximization with Conditionally Independent Label Subsets}

\author{Maxime Gasse%
\and Alex Aussem%
}
\authorrunning{F-measure Maximization with Conditionally Independent Label Subsets}

\institute{LIRIS, UMR 5205,\\
University of Lyon 1, 69622 Lyon, France\\
\mailsa\\
}

\toctitle{F-measure Maximization with Conditionally Independent Label Subsets}
\tocauthor{F-measure Maximization with Conditionally Independent Label Subsets}
\maketitle

\begin{abstract}
We discuss a method to improve the exact F-measure maximization algorithm called GFM, proposed in \autocite{conf/nips/DembczynskiWCH11} for multi-label classification, assuming the label set can be partitioned into conditionally independent subsets given the input features. If the labels were all independent, the estimation of only $m$ parameters ($m$ denoting the number of labels) would suffice to derive Bayes-optimal predictions in $O(m^2)$ operations~\autocite{conf/icml/NanCLC12}. In the general case, $m^2 + 1$ parameters are required by GFM, to solve the problem in  $O(m^3)$ operations. In this work, we show that the number of parameters can be reduced further to $m^2/n$, in the best case, assuming the label set can be partitioned into $n$ conditionally independent subsets. As this label partition needs to be estimated from the data beforehand, we use first the procedure proposed in \autocite{conf/icml/GasseAE15} that finds such partition and then infer the required parameters locally in each label subset. The latter are aggregated and serve as input to GFM to form the Bayes-optimal prediction. We show on a synthetic experiment that the reduction in the number of parameters brings about significant benefits in terms of performance. 
\end{abstract}

\section{Introduction}

Multi-label classification (MLC) has received increasing attention in the last years from the machine learning community. Unlike in the case of multi-class learning, in MLC each instance can be assigned simultaneously to multiple binary labels. Formally, learning from multi-label examples amounts to finding a mapping from a space of features to a space of labels. Given a multi-label training set ${\mathcal D}$, the goal of multi-label learning is to find a function which is able to map any unseen example to its proper set of labels. From a Bayesian point of view, this problem amounts to modeling the conditional joint distribution $p(\mathbf{y} | \mathbf{x})$, where $\mathbf{x}$ is a random vector in $\mathbb{R}^d$ associated with the input space, $\mathbf{y}$ a random vector in $\{0,1\}^m$ associated with the labels, and $p$ the probability distribution defined over $(\mathbf{x},\mathbf{y})$.  Knowing the label conditional distribution $p(\mathbf{y} | \mathbf{x})$ still leaves us with the question of deciding what prediction $\mathbf{y}$ should be made given $\mathbf{x}$ in order to minimize the loss. \textcite{journals/ml/DembczynskiWCH12} show that the expected benefit of exploiting label dependence depends on the type of loss to be minimized and, most importantly, one cannot expect the same MLC method to be optimal for different types of losses at the same time. In particular, optimizing the subset $0/1$ loss, the F-measure loss or the {\em Jaccard} index requires some knowledge of the dependence structure among the labels that cannot be inferred from the marginals $p(y_i | \mathbf{x})$ alone.

The F-measure is a standard performance metric in information retrieval that was used in a variety of prediction problems including binary classification, multi-label classification and structured output prediction. Let $\mathbf{y} = (y_1,\ldots,y_m)$ denote the label vector associated with a single instance $\mathbf{x}$ in MLC, and  $\mathbf{h} = (h_1,\ldots,h_m) \in \{0,1\}^m$ denote the prediction for $\mathbf{x}$, the F-measure is defined as follows:

\begin{equation}
    F(\mathbf{y},\mathbf{h}) = \frac{2 (\mathbf{y} \cdot \mathbf{h})}{\mathbf{y} \cdot \mathbf{y} + \mathbf{h} \cdot \mathbf{h}}
\text{,}
\end{equation}
where $\cdot$ denotes the dot product operator\footnote{In a binary setting the dot product $\mathbf{h} \cdot \mathbf{y}$ offers a convenient notation to count the number of positives values common to both $\mathbf{h}$ and $\mathbf{y}$.} and $0/0=1$ by definition. Optimizing the F-measure is a statistically and computationally challenging problem, since no closed-form solution exists and few theoretical studies of the F-measure were carried out. Very recently, \textcite{journals/jmlr/WaegemanDJCH14} presented a new Bayes-optimal algorithm regardless of the underlying distribution that is statistically consistent. Assuming the underlying probability distribution $p$ is known, the optimal prediction $\mathbf{h}^*$ that maximizes the expected F-measure is given by

\begin{equation}
\label{eq:f-meas-argmax}
    \mathbf{h}^*
    = \argmax_{\mathbf{h} \in \{0,1\}^m} \mathbb{E}_\mathbf{y}[F(\mathbf{y},\mathbf{h})]
    = \argmax_{\mathbf{h} \in \{0,1\}^m} \sum_{\mathbf{y} \in \{0,1\}^m} p(\mathbf{y}) F(\mathbf{y},\mathbf{h})
\text{.}
\end{equation}

The corresponding optimization problem is non-trivial and cannot be solved in closed form. Moreover, a brute-force search is intractable, as it would require checking all $2^m$ combinations of prediction vector $\mathbf{h}$ and summing over an exponential number of terms in each combination. As a result, many works reporting the F-measure in experimental studies rely on optimizing a surrogate loss like the Hamming loss and the subset zero-one loss as an approximation of (\ref{eq:f-meas-argmax}). However, \textcite{journals/jmlr/WaegemanDJCH14} have shown that these surrogate loss functions yield a high {\em worst-case regret}.

Apart from optimizing surrogates, a few other  approaches for finding the F-measure maximizer have been presented but they explicitly rely on the restrictive assumption of independence of the $Y_i$~\autocite{conf/acl/Jansche07,conf/icml/NanCLC12}. This assumption is not tenable in domains like MLC and structured output prediction. Algorithms based on independence assumptions or marginal probabilities are not statistically consistent when arbitrary probability distributions $p$ are considered. 

In \autocite{conf/icml/GasseAE15}, we established several results to characterize and compute disjoint label subsets called {\em irreducible label factors} (ILFs) that appear in the factorization of $p(\mathbf{y} | \mathbf{x})$ (i.e., minimal subsets $\mathbf{Y}_{LF} \subseteq \mathbf{Y}$ such that  $\mathbf{Y}_{LF} \indep \mathbf{Y} \setminus \mathbf{Y}_{LF} \mid \mathbf{X}$) under various assumption underlying the probability distribution. In that paper, the emphasis was placed on the {\em subset zero-one} loss minimization. In the present work, we show that ILF decomposition can also benefit to the F-measure maximization problem in the MLC context.

Section~\ref{sec:gfm} introduces the General F-measure Maximizer method (GFM) from~\autocite{conf/nips/DembczynskiWCH11}. Section~\ref{sec:f-gfm} discusses some key concepts about irreducible label factors, and addresses the problem of exploiting a label factor decomposition within GFM, with an exact procedure called Factorized GFM (F-GFM). Section~\ref{sec:params} presents a practical calibrated parametrization method for GFM and F-GFM, and finally section~\ref{sec:experiments} presents a synthetic experiment to corroborate our theoretical findings.

\section{The General F-measure Maximizer method}
\label{sec:gfm}

We start by reviewing the General F-measure Maximizer method presented in \textcite{conf/nips/DembczynskiWCH11}. \textcite{conf/acl/Jansche07} noticed that (\ref{eq:f-meas-argmax}) can be solved via outer and inner maximization. The inner maximization step is
\begin{equation}
\label{eq:f-meas-argmax-inner}
    \mathbf{h}^{(k)} = \argmax_{\mathbf{h} \in \mathcal{H}_k} \mathbb{E}_\mathbf{y}[F(\mathbf{y},\mathbf{h})]
\text{,}
\end{equation}
where $\mathcal{H}_k = \{\mathbf{h} \in \{0,1\}^m | \mathbf{h} \cdot \mathbf{h} = k\}$, followed by an outer maximization

\begin{equation}
\label{eq:f-meas-argmax-outer}
    \mathbf{h}^* = \argmax_{\mathbf{h} \in \{\mathbf{h}^{(0)},\dots,\mathbf{h}^{(m)}\}} \mathbb{E}_\mathbf{y}[F(\mathbf{y},\mathbf{h})]
\text{.}
\end{equation}

The outer maximization (\ref{eq:f-meas-argmax-outer}) can be done in linear time by simply checking all $m + 1$ possibilities. The main effort is then devoted to solving the inner maximization (\ref{eq:f-meas-argmax-inner}). For convenience, \textcite{journals/jmlr/WaegemanDJCH14} introduce the following quantities:
\[
s_\mathbf{y} = \mathbf{y} \cdot \mathbf{y}
\text{,}\quad
\Delta_{ik} = \sum_{\mathbf{y} \in \mathcal{Y}_i} \frac{2p(\mathbf{y})}{s_\mathbf{y} + k}
\text{,}
\]
with $\mathcal{Y}_i = \{\mathbf{y} \in \{0,1\}^m | y_i = 1\}$. The first quantity is the number of ones in the label vector $\mathbf{y}$, while $\Delta_{ik}$ is a specific marginal value for the $i$-th label. Using these quantities, the maximizer in (\ref{eq:f-meas-argmax-inner}) becomes
\[
\mathbf{h}^{(k)} = \argmax_{\mathbf{h} \in \mathcal{H}_k} \sum_{i=1}^m h_i \Delta_{ik}
\text{,}
\]
which boils down to selecting the $k$ labels with the highest $\Delta_{ik}$ value. In the special case of $k = 0$, we have $\mathbf{h}^{(0)} = 0$ and $\mathbb{E}_\mathbf{y}[F(\mathbf{y},\mathbf{h}^{(0)})] = p(\mathbf{y} = \mathbf{0})$. As a result, it is not required to estimate the $2^m$ parameters of the whole distribution $p(\mathbf{y})$ to find the F-measure maximizer $\mathbf{h}^*$, but only $m^2 + 1$ parameters: the values of $\Delta_{ik}$ which take the form of an $m \times m$ matrix $\mathbf{\Delta}$, plus the value of $p(\mathbf{y} = \mathbf{0})$.

The resulting algorithm is referred to as General F-measure Maximizer (GFM), and yields the optimal F-measure prediction in $O(m^2)$ (see \autocite{journals/jmlr/WaegemanDJCH14} for details). In order to combine GFM with a training algorithm, the authors decompose the $\mathbf{\Delta}$ matrix as follows. Consider the probabilities
\[
p_{is} = p(y_i = 1, s_\mathbf{y} = s)
\text{,}\quad
i,s \in \{1, \dots, m\}
\]
that constitute an $m \times m$ matrix $\mathbf{P}$, along an $m \times m$ matrix $\mathbf{W}$ with elements
\[
w_{sk} = \frac{2}{s + k}
\text{,}
\]
then it can be easily shown that
\begin{equation}
\label{eq:pw-prod}
\mathbf{\Delta} = \mathbf{P}\mathbf{W}
\text{.}
\end{equation}
If the matrix $\mathbf{P}$ is taken as an input by the algorithm, then its complexity is dominated by the matrix multiplication (\ref{eq:pw-prod}), which is solved naively in $O(m^3)$.

In view of this result, \textcite{conf/nips/DembczynskiWCH11} establish that modeling pairwise or higher degree dependences between labels is not necessary to obtain an optimal solution, only a proper estimation of marginal quantities $p_{is}$ is required to take the number of co-occurring labels into account. In this work, we will show that modeling high degree dependences between labels can help to obtain a better estimation of $p_{is}$, and thereby better predictions within the GFM framework.

\section{Factorized GFM}
\label{sec:f-gfm}

In the following we will show that, assuming a factorization of the conditional distribution of the labels, the $p_{is}$ parameters can be reconstructed from a smaller number of parameters that are estimated locally in each label factor, at a computational cost of $O(m^3)$.

\subsection{Label factor decomposition}

We now introduce the concept of {\em label factor} that will play a pivotal role in the factorization of $p(\mathbf{y} | \mathbf{x})$ \autocite{conf/icml/GasseAE15}.

\begin{definition}
\label{def:lf}
A label factor is a label subset $\mathbf{Y}_{F} \subseteq \mathbf{Y}$ such that $\mathbf{Y}_{F} \indep \mathbf{Y} \setminus \mathbf{Y}_{F} \mid \mathbf{X}$. Additionally, a label factor is said irreducible when it is non-empty and has no other non-empty label factor as proper subset.
\end{definition}

The key idea behind irreducible label factors  (ILFs as a shorthand) is the decomposition of the conditional distribution of the labels into a product of factors,
\[
p(\mathbf{y}\mid \mathbf{x}) = \prod_{k=1}^n p(\mathbf{y}_{F_k}  \mid \mathbf{x})
\text{,}
\]
where $\{\mathbf{Y}_{F_k}\}_{k=1}^n$  is a partition of $\mathbf{Y} = \{Y_1,Y_2,\ldots,Y_m\}$. From the above definition, we have that $\mathbf{Y}_{F_i} \indep \mathbf{Y}_{F_j} \mid \mathbf{X}$, $\forall i \neq j$. To illustrate the concept of label factor decomposition, consider the following example.

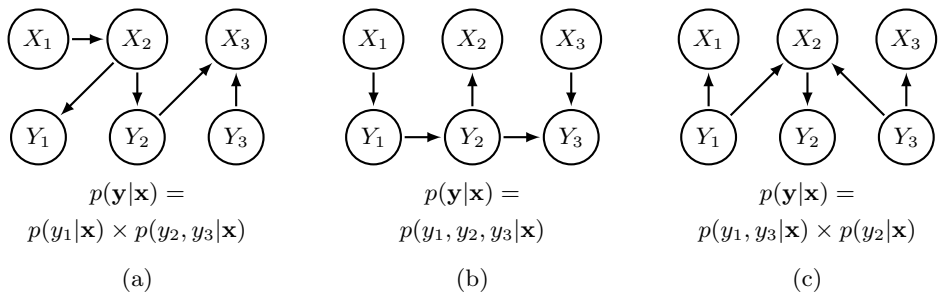
\begin{figure}[htb]
    \centering
    \subfloat[]{
        \centering
        \label{fig:example-dag-1}
        \begin{tikzpicture}
            \tikzstyle{dag.node}=[draw, circle, thick, minimum size=7mm]
            \tikzstyle{dag.edge}=[->, >=latex, thick, shorten >=1pt, shorten <=1pt]
            \node[dag.node] (X1) at (0,1.3) {$X_1$};
            \node[dag.node] (X2) at (1.3,1.3) {$X_2$};
            \node[dag.node] (X3) at (2.6,1.3) {$X_3$};
            \node[dag.node] (Y1) at (0,0) {$Y_1$};
            \node[dag.node] (Y2) at (1.3,0) {$Y_2$};
            \node[dag.node] (Y3) at (2.6,0) {$Y_3$};
            \draw[dag.edge] (X1) -- (X2);
            \draw[dag.edge] (X2) -- (Y2);
            \draw[dag.edge] (X2) -- (Y1);
            \draw[dag.edge] (Y3) -- (X3);
            \draw[dag.edge] (Y2) -- (X3);
            \node[draw=none,fill=none,align=center,text width=12em] at (1.3, -0.8) {
            \begin{gather*}
                p(\mathbf{y} | \mathbf{x}) = \\
                p(y_1 | \mathbf{x}) \times p(y_2,y_3 | \mathbf{x})
            \end{gather*}
            };
        \end{tikzpicture}
    }
    \hspace*{\fill}
    \subfloat[]{
        \centering
        \label{fig:example-dag-2}
        \begin{tikzpicture}
            \tikzstyle{dag.node}=[draw, circle, thick, minimum size=7mm]
            \tikzstyle{dag.edge}=[->, >=latex, thick, shorten >=1pt, shorten <=1pt]
            \node[dag.node] (X1) at (0,1.3) {$X_1$};
            \node[dag.node] (X2) at (1.3,1.3) {$X_2$};
            \node[dag.node] (X3) at (2.6,1.3) {$X_3$};
            \node[dag.node] (Y1) at (0,0) {$Y_1$};
            \node[dag.node] (Y2) at (1.3,0) {$Y_2$};
            \node[dag.node] (Y3) at (2.6,0) {$Y_3$};
            \draw[dag.edge] (X1) -- (Y1);
            \draw[dag.edge] (Y2) -- (X2);
            \draw[dag.edge] (X3) -- (Y3);
            \draw[dag.edge] (Y1) -- (Y2);
            \draw[dag.edge] (Y2) -- (Y3);
            \node[draw=none,fill=none,align=center,text width=12em] at (1.3, -0.8) {
            \begin{gather*}
                p(\mathbf{y} | \mathbf{x}) = \\
                p(y_1,y_2,y_3 | \mathbf{x})
            \end{gather*}
            };
        \end{tikzpicture}
    }
    \hspace*{\fill}
    \subfloat[]{
        \centering
        \label{fig:example-dag-4}
        \begin{tikzpicture}
            \tikzstyle{dag.node}=[draw, circle, thick, minimum size=7mm]
            \tikzstyle{dag.edge}=[->, >=latex, thick, shorten >=1pt, shorten <=1pt]
            \node[dag.node] (X1) at (0,1.3) {$X_1$};
            \node[dag.node] (X2) at (1.3,1.3) {$X_2$};
            \node[dag.node] (X3) at (2.6,1.3) {$X_3$};
            \node[dag.node] (Y1) at (0,0) {$Y_1$};
            \node[dag.node] (Y2) at (1.3,0) {$Y_2$};
            \node[dag.node] (Y3) at (2.6,0) {$Y_3$};
            \draw[dag.edge] (Y1) -- (X1);
            \draw[dag.edge] (X2) -- (Y2);
            \draw[dag.edge] (Y3) -- (X3);
            \draw[dag.edge] (Y1) -- (X2);
            \draw[dag.edge] (Y3) -- (X2);
            \node[draw=none,fill=none,align=center,text width=12em] at (1.3, -0.8) {
            \begin{gather*}
                p(\mathbf{y} | \mathbf{x}) = \\
                p(y_1, y_3 | \mathbf{x}) \times p(y_2 | \mathbf{x})
            \end{gather*}
            };
        \end{tikzpicture}
    }
    \caption{Three Bayesian networks for illustration purposes, along with the induced factorization of $p(\mathbf{y} | \mathbf{x})$.}
    \label{fig:example-dags}
\end{figure}

\begin{example}
Suppose $p$ is faithful to one of the DAGs displayed in Figure~\ref{fig:example-dags}.
In DAG~\ref{fig:example-dag-1}, it is easily shown using the $d$-separation criterion that $\{Y_1\} \indep \{Y_2,Y_3\} \mid \mathbf{X}$, so both $\{Y_1\}$ and $\{Y_2,Y_3\}$ are label factors. However, we have $\{Y_2\} \not\indep \{Y_1,Y_3\} \mid \mathbf{X}$ and $\{Y_3\} \not\indep \{Y_1,Y_2\} \mid \mathbf{X}$, so $\{Y_2\}$ and $\{Y_3\}$ are not label factors. Therefore $\{Y_1\}$ and $\{Y_2,Y_3\}$ are the only irreducible label factors. Likewise, in DAG~\ref{fig:example-dag-2} the only irreducible label factor is $\{Y_1,Y_2,Y_3\}$. Finally, in DAG~\ref{fig:example-dag-4} we have that $\{Y_1\} \not\indep \{Y_2,Y_3\} \mid \mathbf{X}$, $\{Y_2\} \indep \{Y_1,Y_3\} \mid \mathbf{X}$ and $\{Y_3\} \not\indep \{Y_1,Y_2\} \mid \mathbf{X}$, so $\{Y_2\}$ and $\{Y_1,Y_3\}$ are the irreducible label factors.
 \end{example}

For convenience, the conditioning on $\mathbf{X}$ will be made implicit in the remainder of this work. Let $m_k$ denote the number of labels in a particular label factor, we introduce for every label factor $\mathbf{Y}_{F_k}=\{Y_1,\dots,Y_{m_k}\}$ the following terms,
\begin{gather*}
p_{is}^{k} = p(y_i=1,s_{\mathbf{y}_{F_k}}=s)
\text{,} \quad i,s \in \{1, \dots, m_k\}
\text{,}
\end{gather*}
which constitute an $m_k \times m_k$ matrix $\mathbf{P}^k$.

Given a factorization of the label set into label factors, our proposed method called F-GFM requires to estimate, for each label factor, a local matrix $\mathbf{P}^{k}$ of size ${m_k}^2$, and then combine them to reconstruct the global matrix $\mathbf{P}$ of size $m^2$.  The total number of parameters is therefore reduced from $m^2$ to $\sum_{k=1}^{n} {m_k}^2$. It is easily shown that, in the best case, the total number of parameters is ${m^2}/n$ when $m_k=m/n$ for every label factor, and the worst case is $(n-1)+(m-n+1)^2$ when all the label factors, but one, are singletons. In both cases the number of parameters is reduced, which  results in a better estimation of these parameters and a better robustness of the model. In the following, we describe a procedure to recover $\mathbf{P}$ and $p(\mathbf{y}=\mathbf{0})$ from the individual $\mathbf{P}^k$ matrices in $O(m^3)$.

\subsection{Recovering \texorpdfstring{$\mathbf{d}^k$}{dk}}

Consider, for every label factor $\mathbf{Y}_{F_k}$, the following probabilities,
\[
d_{s}^{k} = p(s_{\mathbf{y}_{F_k}}=s)
\text{,} \quad
s \in \{0, \dots, m_k\}
\text{,}
\]
which form a vector $\mathbf{d}^{k}$ of size $m_k+1$. Instead of estimating these additional terms, they are extracted directly from $\mathbf{P}^k$ in ${m_k}^2$ operations. Extracting these parameters is done prior to recovering $\mathbf{P}$. We now describe how to recover a particular $\mathbf{d}^k$ vector from a $\mathbf{P}^k$ matrix. Note that the same method holds to recover $\mathbf{d}$ from $\mathbf{P}$, therefore in the following we will drop the superscript $k$  to keep our notations uncluttered. Consider the following expression for $p_{is}$ and $d_s$,
\begin{align*}
p_{is} &= \sum_{\mathbf{y} \in \{0,1\}^m} p(\mathbf{y}) \cdot \mathbb{I}[s_\mathbf{y}=s] \cdot \mathbb{I}[y_i=1]
\text{,} \\
d_{s} &= \sum_{\mathbf{y} \in \{0,1\}^m} p(\mathbf{y}) \cdot \mathbb{I}[s_\mathbf{y}=s]
\text{.}
\end{align*}
Notice that, for a particular $\mathbf{y} \in \{0,1\}^m$, the following equality holds,
\[
\mathbb{I}[s_{\mathbf{y}}=s] \cdot \sum_{i=1}^{m} \mathbb{I}[y_i=1]  = s \cdot \mathbb{I}[s_{\mathbf{y}}=s]
\text{.}
\]
Therefore, when $s>0$, $d_{s}$ can be expressed as
\[
d_{s} = \sum_{\mathbf{y} \in \{0,1\}^m} p(\mathbf{y}) \cdot \mathbb{I}[s_\mathbf{y}=s] \cdot \frac{1}{s} \sum_{i=1}^{m} \mathbb{I}[y_i=1]
\text{.}
\]
This expression can be further simplified in order to express $d_s$ as a composition of $p_{is}$ terms,
\begin{align*}
d_{s} = \frac{1}{s} \sum_{i=1}^{m} p_{is} \text{,} \quad \forall
 s \in \{1, \dots, m\}
\text{.}
\end{align*}
We may recover $d_{0}$ from
\[
d_{0} = 1 - \sum_{s=1}^{m} d_{s}
\text{.}
\]
As a result, each vector $\mathbf{d}^k$  can be obtained from $\mathbf{P}^k$ in ${m_k}^2$ operations. Interestingly, because $p(\mathbf{y}=\mathbf{0})=d_{0}$, this additional parameter can actually be inferred from $\mathbf{P}$ at the expense of $m^2$ operations, thereby reducing the number of parameters required by GFM to $m^2$ instead of $m^2+1$.

\subsection{Recovering \texorpdfstring{$\mathbf{P}$}{P}}

We will now show how the whole $\mathbf{P}$ matrix can be recovered from the $\mathbf{P}^k$ matrices in $O(m^3)$.

\subsubsection{When \texorpdfstring{$n=2$}{n=2}.} Let us first assume that there are only two label factors $\mathbf{Y}_{F_1}$ and $\mathbf{Y}_{F_2}$. Consider a label $Y_i$ that belongs to $\mathbf{Y}_{F_1}$, from the marginalization rule $p_{is}$ may be decomposed as follows,
\begin{align}
\label{eq:pis-1}
p_{is} = \sum_{s'} p(y_i=1,s_\mathbf{y}=s,s_{\mathbf{y}_{F_1}}=s')
\text{.}
\end{align}
The inner term of this sum factorizes because of the label factor assumption. First, recall that $s_\mathbf{y}=s_{\mathbf{y}_{F_1}}+s_{\mathbf{y}_{F_2}}$, which allows us to write
\[
p(y_i=1, s_\mathbf{y}=s, s_{\mathbf{y}_{F_1}}=s') = p(y_i=1, s_{\mathbf{y}_{F_1}}=s', s_{\mathbf{y}_{F_2}}=s-s')
\text{.}
\]
Second, due to the label factor assumption, i.e. $\mathbf{Y}_{F_1} \indep \mathbf{Y}_{F_2}$, we have
\begin{align}
\label{eq:pis-2}
p(y_i, s_\mathbf{y}, s_{\mathbf{y}_{F_1}}) = p(y_i, s_{\mathbf{y}_{F_1}}) \cdot p(s_{\mathbf{y}_{F_2}})
\text{.}
\end{align}
We may combine (\ref{eq:pis-2}) and (\ref{eq:pis-1}) to obtain

\begin{align}
\label{eq:pis-3}
p_{is} = \sum_{s'} p(y_i=1, s_{\mathbf{y}_{F_1}}=s') \cdot p(s_{\mathbf{y}_{F_2}}=s-s')
\text{.}
\end{align}
Finally, we have necessarily $s' \leq s$ and $s' \leq m_1$, which implies $s' \leq min(s, m_1)$. Also, $s-s' \leq m_2$ and $s' \geq 1$ because $y_i=1$, which implies $s' \geq max(1, s-m_2)$. So we can re-write (\ref{eq:pis-3}) as follows,
\begin{align}
\label{eq:pis-4}
p_{is} = \sum_{s'= max(1, s-m_2)}^{min(s, m_1)} p_{is'}^{1} \cdot d_{s-s'}^{2}
\text{.}
\end{align}

In the case where $Y_i \in \mathbf{Y}_{F_2}$, we obtain a similar result. In the end, given that both $\mathbf{P}^{k}$ and $\mathbf{d}^{k}$ are known for $\mathbf{Y}_{F_1}$ and $\mathbf{Y}_{F_2}$, (\ref{eq:pis-4}) allows us to recover all term in $\mathbf{P}$ in $(m_2+1){m_1}^2 + (m_1+1){m_2}^2$ operations. Assuming that only the $\mathbf{P}^k$ matrices are known, we must add up the additional cost for recovering the $\mathbf{d}^k$ vectors, which brings the total computational burden to $(m_2+2){m_1}^2 + (m_1+2){m_2}^2$.

\subsubsection{For any \texorpdfstring{$n$}{n}.} The same procedure can be used iteratively to merge $\mathbf{P}^1$ and $\mathbf{P}^2$ into a matrix $\mathbf{P}'$ of size $(m_1+m_2)^2$, then combine this  matrix with $\mathbf{P}^3$ to form a new matrix of size $(m_1+m_2+m_3)^2$, and so on until every label factor is merged into a matrix of size $m^2$. In the end we obtain $\mathbf{P}$ in a total number of operations equal to
\[
\sum_{i=2}^{n}(m_i+2)(\sum_{j=1}^{i-1}m_j)^2+m_i^2(2+\sum_{j=1}^{i-1}m_j) \text{.}
\]
To avoid tedious calculations, we can easily compute a tight upper bound of the number of computations, i.e.
\[
\max_{m_1,\dots,m_n} \quad \sum_{i=2}^{n}(m_i+2)\left(\sum_{j=1}^{i-1}(m_j+2)\right)\left(\sum_{j=1}^{i}(m_j+2)\right) \quad \text{s.t.} \sum_{i=1}^{n} m_i = m \text{.}
\]
Solving $\nabla\Lagr(m_1,\dots,m_n,\lambda)=0$ yields
\[
m_i = \left((m+2n)^2 - \lambda\right)^{1/2} + 2n \text{,} \quad \forall i \in \{1, \dots, n\} \text{,}
\]
which implies that all the label factors have equal size. As a result, with $m_i = m/n$ for every label factor we obtain an upper bound on the worst case number of operations equal to $(\frac{m}{n}+2)^3(n^2-1)$. Thus, the overall complexity to recover $\mathbf{P}$ is bounded by $O(m^3)$.

\subsection{The F-GFM algorithm}

Given that the label factors are known and that every $\mathbf{P}^k$ matrix has been estimated, the whole procedure for recovering $\mathbf{P}$ and then $\mathbf{h}^*$ is presented in Algorithm~\ref{alg:f-gfm}. As shown in the previous section, the overall complexity of F-GFM is $O(m^3)$, just as GFM.

\begin{algorithm}[htb]
  \caption{Factorized-GFM}
  \label{alg:f-gfm}
  \begin{algorithmic}[1]
    \Require $\mathbf{Y}$ the label set, $\mathbf{Y}_{F_1}, \dots, \mathbf{Y}_{F_n}$ the label factors, $m_1, \dots, m_n$ their size and $\mathbf{P}^1, \dots, \mathbf{P}^n$ their matrix of $p_{i,s}^k$ parameters.
    \Ensure $\mathbf{h}^*$ the F-measure maximizing prediction.
    
    \State Initialize $m \gets 0$, $\mathbf{P} \gets \emptyset$, $\mathbf{d} \gets \{1\}$
    \ForAll{$k \in \{1, \dots, n\}$}
        \State $m' \gets m$, $\mathbf{P}' \gets \mathbf{P}$, $\mathbf{d}' \gets \mathbf{d}$, $m \gets m' + m_k$
        
        \State Initialize $\mathbf{d}^k = \{d_0, \dots, d_{m_k}\}$ a vector of size $m_k+1$
        \ForAll{$s \in \{1, \dots, m_k\}$} \Comment{1) recover $\mathbf{d}^k$ from $\mathbf{P}^k$}
            \State $d_s^k \gets s^{-1} \sum_{i=1}^{m_k} p_{i,s}^k$
        \EndFor
        \State $d_0^k \gets 1 - \sum_{s=1}^{m_k} d_s^k$
        
        \State Initialize $\mathbf{P}$ a zero matrix of size $m \times m$
        \ForAll{$i \in \{1, \dots, m_k\}$} \Comment{2) merge $\mathbf{P}^k$ and $\mathbf{d}'$ into $\mathbf{P}$}
            \ForAll{$s1 \in \{1, \dots, m_k\}$}
                \ForAll{$s2 \in \{0, \dots, m'\}$}
                    \State $p_{i,s1+s2} \gets p_{i,s1+s2} + p_{i,s1}^k \cdot d'_{s2}$
                \EndFor
            \EndFor
        \EndFor
        \ForAll{$i \in \{1, \dots, m'\}$} \Comment{3) merge $\mathbf{P}'$ and $\mathbf{d}^k$ into $\mathbf{P}$}
            \ForAll{$s1 \in \{1, \dots, m'\}$}
                \ForAll{$s2 \in \{0, \dots, m_k\}$}
                    \State $p_{i+m_k,s1+s2} \gets p_{i+m_k,s1+s2} + p'_{i,s1} \cdot d_{s2}^k$
                \EndFor
            \EndFor
        \EndFor
        
        \State Initialize $\mathbf{d}$ a zero vector of size $m+1$
        \ForAll{$s \in \{1, \dots, m\}$} \Comment{4) recover $\mathbf{d}$ from $\mathbf{P}$}
            \State $d_s \gets s^{-1} \sum_{i=1}^{m} p_{i,s}$
        \EndFor
        \State $d_0 \gets 1 - \sum_{s=1}^{m} d_s$
    \EndFor
    
    \State $\mathbf{h}^* \gets GFM(\mathbf{P}, d_0)$ \Comment{5) obtain $\mathbf{h}^*$ from $\mathbf{P}$ and $d_0$}
    \State Rearrange $\mathbf{h}^*$ to match the order of the labels in $\mathbf{Y}$.
  \end{algorithmic}
\end{algorithm}

\section{Parameter estimation}
\label{sec:params}

Our proposed method F-GFM requires to estimate for each label factor $\mathbf{Y}_{F_k}$ the $m_k \times m_k$ matrix $\mathbf{P}^{k}$, instead of the whole $m \times m$ matrix $\mathbf{P}$ in GFM. Still, the problem of parameter estimation in GFM and F-GFM is essentially the same, that is, estimating the matrix $\mathbf{P}$ (resp. $\mathbf{P}^{k}$) for a particular input $\mathbf{x}$, given a set of training samples $(\mathbf{x},\mathbf{y})$ (resp. $(\mathbf{x},\mathbf{y}_{F_k})$).

\textcite{conf/icml/DembczynskiJKWH13} propose a solution to estimate the $p_{is}$ terms directly, by solving $m$ multinomial logistic regression problems with $m+1$ classes. For each label $Y_i$ the scheme of the reduction is the following:
\[
(\mathbf{x},\mathbf{y}) \rightarrow (\mathbf{x}, y = y_i \cdot s_{\mathbf{y}})
\text{.}
\]
However, we observed that the parameters estimated with this approach are inconsistent, that is, they often result in a negative probability for $d_0$ when trying to recover $\mathbf{d}$ from $\mathbf{P}$. To overcome this numerical problem, we found a straightforward and effective approach. Instead of estimating the $p_{is}$ terms directly, we can proceed in two steps. From the chain rule of probabilities, we have that
\begin{align}
\label{eq:params}
p(y_i, s_\mathbf{y}|\mathbf{x}) = p(s_\mathbf{y}|\mathbf{x}) \cdot p(y_i|s_\mathbf{y},\mathbf{x})
\text{.}
\end{align}
The idea is to estimate each of these two terms independently. First, the $p(s_\mathbf{y}=s|\mathbf{x})$ terms are obtained by performing multinomial logistic regression with $m+1$ classes, using the following mapping:
\[
(\mathbf{x},\mathbf{y}) \rightarrow (\mathbf{x}, y = s_{\mathbf{y}})
\text{.}
\]
Second, for each label $Y_i$ we estimate the $p(y_i=1|s_\mathbf{y}=s,\mathbf{x})$ terms with a  binary logistic regression model, using the following mapping:
\[
(\mathbf{x},\mathbf{y}) \rightarrow ((\mathbf{x}, s_{\mathbf{y}}), y = y_i )
\text{.}
\]
To summarize, for each label factor, one multinomial logistic regression model with $m_k+1$ classes, and $m_k$ binary logistic regression models are trained. In order to estimate the $p_{is}^k$ terms, we combine the outputs of the multinomial and the binary models according to (\ref{eq:params}). This approach has the desirable advantage of producing calibrated $\mathbf{P}^k$ matrices and $\mathbf{d}^k$ vectors, which appears to be crucial  for the success of F-GFM. Notice that in our experiments this approach was also very beneficial to GFM in terms of MLC performance.

\section{Experiments}
\label{sec:experiments}

In this section, we compare GFM and F-GFM on a synthetic toy problem to assess the effective improvement in classification performance due to the label factorization. The code to reproduce this experiment was made available online\footnote{\url{https://github.com/gasse/fgfm-toy}}.

\subsection{Setup details}

\begin{figure}[htb]
    \label{fig:toy-dag}
    \centering
    \begin{tikzpicture}[
            declare function={l=1.5;},
            declare function={m=2;}
        ]
        \tikzstyle{d.node}=[draw, circle, thick, minimum size=7mm]
        \tikzstyle{d.edge}=[->, >=latex, thick, shorten >=1pt, shorten <=1pt]
        \node[draw, ellipse, thick, minimum height=7mm, align=center] (X) at (3.5*l,1*m) {$X_{1,2,3,4}$};
        \node[d.node] (Z1) at (5.5*l,1*m) {$X_5$};
        \node[d.node] (Z2) at (6.5*l,1*m) {$X_6$};
        \node[d.node] (Y1) at (0*l,0*m) {$Y_1$};
        \node[d.node] (Y2) at (1*l,0*m) {$Y_2$};
        \node[d.node] (Y3) at (2*l,0*m) {$Y_3$};
        \node[d.node] (Y4) at (3*l,0*m) {$Y_4$};
        \node[d.node] (Y5) at (4*l,0*m) {$Y_5$};
        \node[d.node] (Y6) at (5*l,0*m) {$Y_6$};
        \node[d.node] (Y7) at (6*l,0*m) {$Y_7$};
        \node[d.node] (Y8) at (7*l,0*m) {$Y_8$};
        
        \draw [d.edge] (X) -- (Y1);
        \draw [d.edge] (X) -- (Y2);
        \draw [d.edge] (X) -- (Y3);
        \draw [d.edge] (X) -- (Y4);
        \draw [d.edge] (X) -- (Y5);
        \draw [d.edge] (X) -- (Y6);
        \draw [d.edge] (X) -- (Y7);
        \draw [d.edge] (X) -- (Y8);
        
        \draw [d.edge] (Y1) -- (Y2);
        \draw [d.edge] (Y1) to[out=-35, in=-145] (Y3);
        \draw [d.edge] (Y1) to[out=-45, in=-135] (Y4);
        \draw [d.edge] (Y2) -- (Y3);
        \draw [d.edge] (Y2) to[out=-35, in=-145] (Y4);
        \draw [d.edge] (Y3) -- (Y4);
        
        \draw [d.edge] (Y5) -- (Y6);
        \draw [d.edge] (Y5) to[out=-35, in=-145] (Y7);
        \draw [d.edge] (Y5) to[out=-45, in=-135] (Y8);
        \draw [d.edge] (Y6) -- (Y7);
        \draw [d.edge] (Y6) to[out=-35, in=-145] (Y8);
        \draw [d.edge] (Y7) -- (Y8);
    \end{tikzpicture}
    \caption{BN structure of our toy problem with DAG 2, i.e. two label factors $\{Y_1,Y_2,Y_3,Y_4\}$ and $\{Y_5,Y_6, Y_7,Y_8\}$. Note that nodes $X_1$, $X_2$, $X_3$ and $X_4$ are grouped up for readability.}
\end{figure}
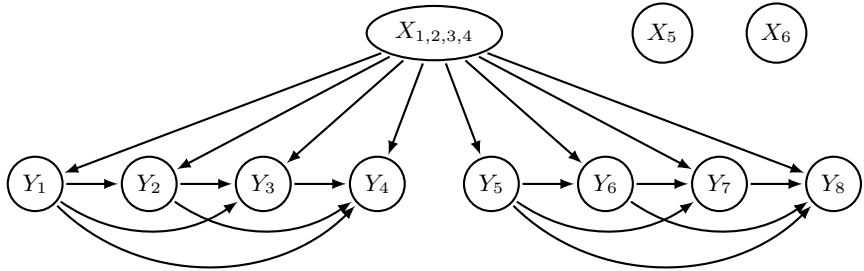

Consider $\mathbf{Y}=\{Y_1, \ldots, Y_8\}$ 8 labels and $\mathbf{X}=\{X_1, \dots, X_6\}$ 6 binary random variables. The true joint distribution $p(\mathbf{x},\mathbf{y})$ is encoded in a Bayesian network (one example is displayed in Fig. \ref{fig:toy-dag}) which imposes different label factor decompositions and serves as a data-generative model. In this BN structure (a directed acyclic graph, DAG for short), each of the features $X_1, X_2, X_3, X_4$ is a parent node to every label, to enable a relationship between $\mathbf{X}$ and $\mathbf{Y}$. The remaining features $X_5$ and $X_6$ are totally disconnected in the graph, and thus serve as irrelevant features. Each label factor $\mathbf{Y}_{F_k}$ is made fully connected by placing an edge $Y_i \rightarrow Y_j$ for every $Y_i,Y_j \in \mathbf{Y}_{F_k}$, $i<j$. As a result each label factor is conditionally independent of the other labels given $\mathbf{X}$, yet it exhibits conditional dependencies between its own labels. We consider 4 distinct structures encoding the following label factor decompositions:
\begin{itemize}
    \item DAG 1: $\{Y_1, Y_2\}, \{Y_3, Y_4\}, \{Y_5, Y_6\}, \{Y_7, Y_8\}$;
    \item DAG 2: $\{Y_1, Y_2, Y_3, Y_4\}, \{Y_5, Y_6, Y_7, Y_8\}$;
    \item DAG 3: $\{Y_1, Y_2, Y_3, Y_4, Y_5, Y_6\}, \{Y_7, Y_8\}$;
    \item DAG 4: $\{Y_1, Y_2, Y_3, Y_4, Y_5, Y_6, Y_7, Y_8\}$.
\end{itemize}

Once these BN structures are fixed, the next step is to generate random distributions $p(\mathbf{x},\mathbf{y})$ to sample from. For each BN structure we generate a probability distribution by sampling uniformly the conditional probability table of each node given its parents, $p(x|\mathbf{pa}_x)$, from a unit simplex as discussed in \textcite{Smith04}. The process is repeated 100 times randomly, and each time we generate 7 data sets with 50, 100, 200, 500, 1000, 2000 and 5000 training samples, and 5000 test samples. We report the comparative performance of GFM and F-GFM on the test samples with respect to each scenario (DAG structure) and each training size, averaged over the 100 repetitions.

\begin{figure}[p]
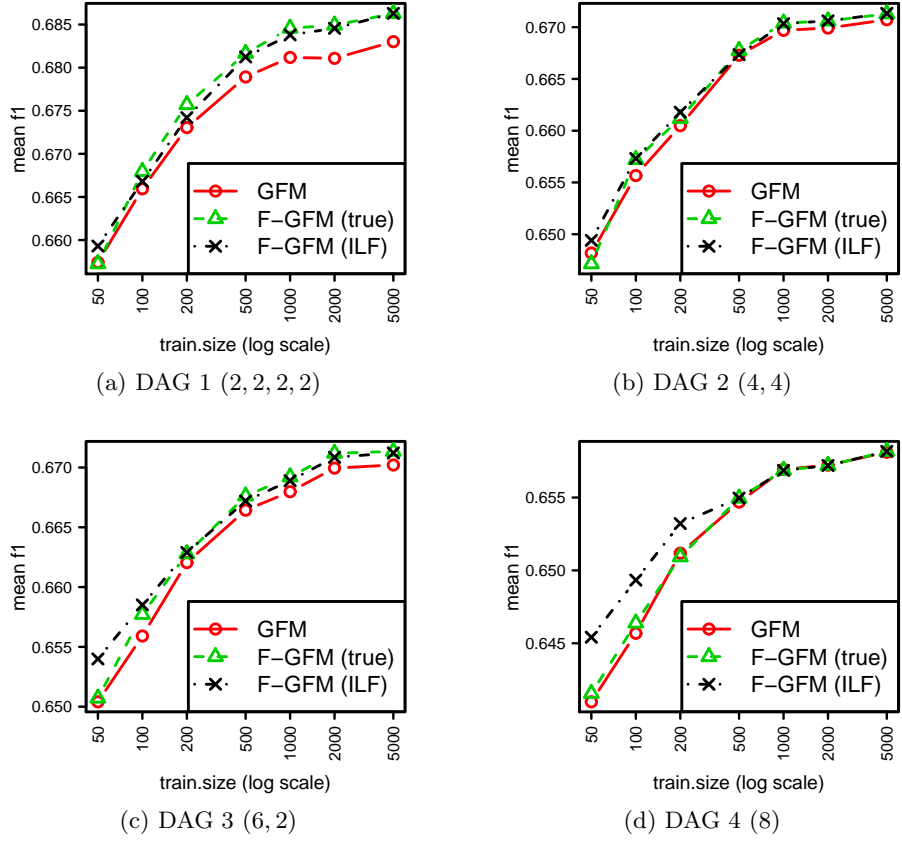

    \centering
    \subfloat[DAG 1 ($2, 2, 2, 2$)]{
        \centering
        \includegraphics[trim=0 0 0.2in 0.2in, clip=true, width=0.45\textwidth]{{{loss.inc.prop.log.4x2.2z2.8y2.bn.lp2.seed0.reps0100.test5000.f1.all}}}
        \label{fig:toy-res1}
    }
    \hspace*{\fill}
    \subfloat[DAG 2 ($4, 4$)]{
        \centering
        \includegraphics[trim=0 0 0.2in 0.2in, clip=true, width=0.45\textwidth]{{{loss.inc.prop.log.4x2.2z2.8y2.bn.lp4.seed0.reps0100.test5000.f1.all}}}
        \label{fig:toy-res2}
    }
    \\
    \subfloat[DAG 3 ($6, 2$)]{
        \centering
        \includegraphics[trim=0 0 0.2in 0.2in, clip=true, width=0.45\textwidth]{{{loss.inc.prop.log.4x2.2z2.8y2.bn.lp6.seed0.reps0100.test5000.f1.all}}}
        \label{fig:toy-res3}
    }
    \hspace*{\fill}
    \subfloat[DAG 4 ($8$)]{
        \centering
        \includegraphics[trim=0 0 0.2in 0.2in, clip=true, width=0.45\textwidth]{{{loss.inc.prop.log.4x2.2z2.8y2.bn.lp8.seed0.reps0100.test5000.f1.all}}}
        \label{fig:toy-res4}
    }
    \caption{Mean $F$-measure of GFM and F-GFM on each DAG, with 7 different training sizes (50, 100, 200, 500, 1000, 2000, 5000) displayed on a logarithmic scale, averaged over 100 repetitions with random distributions. F-GFM (true) uses the true decomposition, while F-GFM (ILF) uses the decomposition learned with ILF-Compo from the training data.}
    \label{fig:toy-res}
\end{figure}

\subsection{Implementation details}

To extract the irreducible label factors, we employ the ILF-Compo algorithm proposed in \autocite{conf/icml/GasseAE15}, with $\alpha=0.01$. To estimate the parameters, we use the standard multinomial logistic regression model from the \textit{nnet}\autocite{book/VenablesR02} R package, with weight decay regularization and $\lambda$ chosen over a 3-fold cross validation.

\subsection{Results}

The comparative performance results for GFM and F-GFM are displayed in Figure~\ref{fig:toy-res} in terms of mean F-measure on the test set averaged over 100 runs, each time using a new probability instantiation. In order to assess separately the influence of the F-GFM procedure and the label factors discovery procedure ILF-Compo, we present two instantiations of F-GFM: one which uses the true decomposition that can be read from the DAG (true), and one obtained from ILF-Compo based on the training data (ILF).

As expected, the more date available for training, the more accurate the parameter estimates, and thus the better the mean F-measure on the test set. F-GFM based on ILF-compo outperforms the original GFM method, sometimes by a significant margin (see Fig.\ref{fig:toy-res3} and \ref{fig:toy-res4} with small sample sizes). Interestingly, F-GFM based on ILF performs not only better than GFM, but also better than F-GFM based on the true label factor decomposition, especially in the last case with a single ILF of size 8 with small sample sizes. The reason is that the label conditional independencies extracted by ILF-Compo are actually observed in the small training sets while being false in the true distribution. As these false label conditional independencies are found almost valid in these small samples - at least from a  numerical point of view - they are exploited by F-GFM to reduce the number of parameters. This is not surprising as Binary Relevance is sometimes shown to outperform other sophisticated MLC techniques exploiting the label correlations while being based on wrong assumptions when training data are insufficient~\autocite{journals/pai/LuacesDBCB12}. The same remark holds for the Naive Bayes model in standard multi-class learning tasks, which wrongly assumes the features to be independent given the output. It is also worth noting that F-GFM with the learned ILF decomposition behaves usually as good or better than F-GFM based on the ground truth ILF decomposition.

\section{Conclusion}

We discussed a method to improve the exact F-measure maximization algorithm (GFM), for multi-label classification, assuming the label set can be partitioned into conditionally independent subsets given the input features. In the general case, $m^2 + 1$ parameters are required by GFM, to solve the problem in $O(m^3)$ operations. In this work, we show that the number of parameters can be reduced further to $m^2/n$, in the best case, assuming the label set can be partitioned into $n$ conditionally independent subsets. As the label partition needs to be estimated from the data beforehand, we use first the procedure proposed in \autocite{conf/icml/GasseAE15} that finds such partition and then infer the required parameters locally in each label subset. The latter are aggregated and serve as input to GFM to form the Bayes-optimal prediction. Our experimental results on a synthetic problem exhibiting various forms of label inpedendencies demonstrate noticeable improvements in terms of F-measure over the standard GFM approach. Interestingly, F-GFM was shown to take advantage of purely fortuitous label independencies in small training sets, despite being false in the underlying distribution, to reduce further the number of parameters, while performing better than F-GFM based on the true decomposition. This is not surprising as Binary Relevance is sometimes shown to outperform other sophisticated MLC techniques exploiting the label correlations while being based on wrong assumptions when training data are insufficient~\autocite{journals/pai/LuacesDBCB12}. Future work will be aimed at reducing further the number of parameters and the overall complexity of the inference algorithm. Large real-world MLC problems will also be considered in the future. 

\section*{Acknowledgements}

This work was funded by both the French state trough the Nano 2017 investment program and the European Community through the European Nanoelectronics Initiative Advisory Council (ENIAC Joint Undertaking), under grant agreement no 324271 (ENI.237.1.B2013).

\printbibliography
\end{document}